\documentclass[sigconf]{acmart}

\AtBeginDocument{%
  }

\setcopyright{acmcopyright}
\copyrightyear{2023}
\acmYear{2023}
\acmDOI{XXXXXXX.XXXXXXX}

\acmConference[Preprint]{}{May}{2023}
  
\acmPrice{15.00}
\acmISBN{978-1-4503-XXXX-X/18/06}

\usepackage{diagbox}
\usepackage{multirow}

\acmSubmissionID{1449}

\begin{document}

\title{Language-Guided 3D Object Detection in Point Cloud for Autonomous Driving}

\author{Wenhao Cheng}
\authornote{Equal Contribution}
\affiliation{
    \institution{Beijing Institute of Technology}
    \country{Beijing, China}}

\author{Junbo Yin*}
\affiliation{
    \institution{Beijing Institute of Technology}
    \country{Beijing, China}}

\author{Wei Li}
\affiliation{\institution{Inceptio}
    \country{Shanghai, China}}

\author{Ruigang Yang}
\affiliation{\institution{Inceptio}
    \country{Shanghai, China}}

\author{Jianbing Shen}
\authornote{Corresponding author, shenjianbingcg@gmail.com}
\affiliation{\institution{SKL-IOTSC, CIS, University of Macau}
    \country{Macau, China}}


\renewcommand{\shortauthors}{W. Cheng et al.}

\begin{abstract}
This paper addresses the problem of 3D referring expression comprehension (REC) in autonomous driving scenario, which aims to ground a natural language to the targeted region in LiDAR point clouds. Previous approaches for REC usually focus on the 2D or 3D-indoor domain, which is not suitable for accurately predicting the location of the queried 3D region in an autonomous driving scene. In addition, the upper-bound limitation and the heavy computation cost motivate us to explore a better solution. In this work, we propose a new multi-modal visual grounding task, termed LiDAR Grounding. Then we devise a Multi-modal Single Shot Grounding (MSSG) approach with an effective token fusion strategy. It jointly learns the LiDAR-based object detector with the language features and predicts the targeted region directly from the detector without any post-processing. Moreover, the image feature can be flexibly integrated into our approach to provide rich texture and color information.  The cross-modal learning enforces the detector to concentrate on important regions in the point cloud by considering the informative language expressions, thus leading to much better accuracy and efficiency. Extensive experiments on the Talk2Car dataset demonstrate the effectiveness of the proposed methods. Our work offers a deeper insight into the LiDAR-based grounding task and we expect it presents a promising direction for the autonomous driving community.
\end{abstract}




\settopmatter{printacmref=false} 
\renewcommand\footnotetextcopyrightpermission[1]{}

\maketitle

\section{Introduction}
The referring expression comprehension (REC)~\cite{wang2019neighbourhood,Yang_2019_CVPR,Yu_2018_CVPR,wang2018learning,Yang_2019_ICCV,Song_2021_CVPR} task has gained increasing attention recently. The goal of REC is to localize a targeted region in a visual scene described by a natural language expression. Meanwhile, great advances have been made in autonomous driving~\cite{geiger2013vision,sun2020scalability,Caesar_2020_CVPR,yurtsever2020survey}. An intuitive idea is to explore the application of REC in autonomous driving scenarios. In this circumstance, a natural language instruction or command can be sent to the self-driving car to be understood and executed (\textit{e.g.}, park the car next to the pedestrian on the right). This is of great significance to human-car interaction and makes the autonomous driving system more user-friendly. To achieve this, an essential step is to enable the self-driving car to accurately localize the referred region in a 3D LiDAR point cloud scene.

\begin{figure}[t]
  \centering
  \includegraphics[scale=0.34]{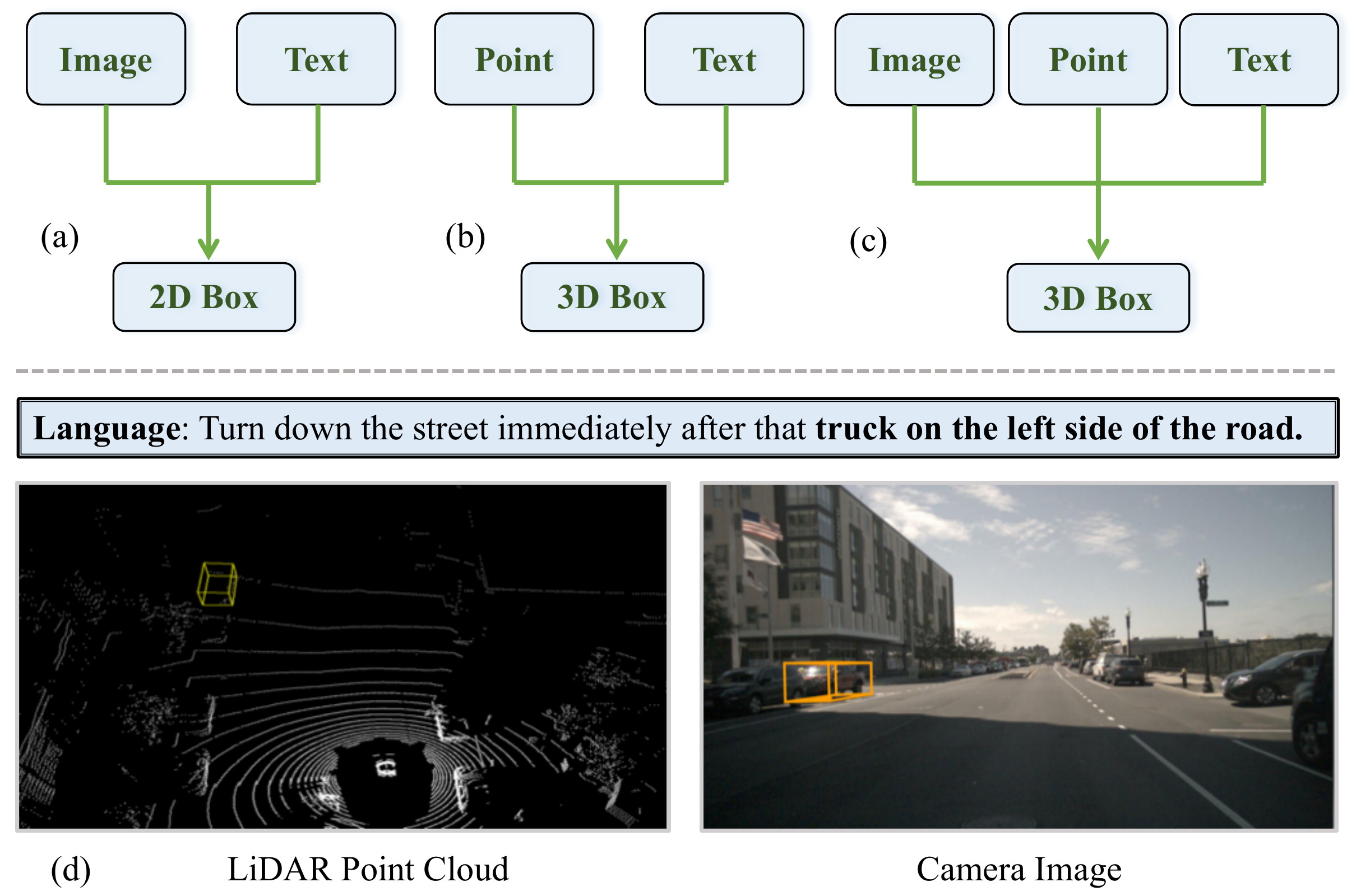}
  \caption{Compared to the previous paradigms like (a) and (b), we propose a new multi-modal setting (c) for  visual grounding in autonomous driving. An example is given in (d), which aims to ground the language referred truck in a LiDAR point cloud scene. The camera image could be optional input to provide context-rich information.}
  \label{fig:shou_tu}
\end{figure}

One line literature in outdoor scene, as shown in Figure~\ref{fig:shou_tu} (a),  Talk2Car~\cite{deruyttere2019talk2car} is the pioneering work in this field. It proposed the first REC dataset in the autonomous driving setting based on nuScenes~\cite{Caesar_2020_CVPR}, which contains commands annotated in natural language expression for self-driving cars. However, they only focus on REC in 2D images~\cite{vandenhende2020baseline}, \textit{i.e.}, grounding the language to a region represented by 2D bounding box. This is not enough in autonomous driving scenarios, since a self-driving car needs to acquire accurate 3D location of the targeted instance in order to execute the given command. Another line of work, as shown in Figure~\ref{fig:shou_tu} (b), such as Referit3D~\cite{achlioptas2020referit3d} and ScanRefer~\cite{chen2020scanrefer}, focus on localizing 3D objects in RGB-D scans using natural language. These methods are infeasible to adapt to outdoor 3D environment. Firstly, the large-scale LiDAR point clouds are sparse, noisy, and lack of informative appearance. How to incorporate the language expression with point cloud representation remains under-explored. Secondly, compared with indoor domain, there are fewer common categories in outdoor scene (car, bicycle, and pedestrian), which increases the difficulty of extracting distinguishable features from natural language.  Moreover, different from the 3D object detection task that handles objects with pre-defined classes, the LiDAR grounding task needs to tackle various instances in different scenes. 


To this end, we propose a new 3D REC task, \textit{i.e.}, LiDAR Grounding, to ground command in natural language to a targeted region described by a 3D bounding box. The paradigm is illustrated in Figure~\ref{fig:shou_tu} (c). To address the aforementioned challenges, we firstly devise an intuitive baseline following Ground-By-Detection (GBD) manner. More precisely, we can apply off-the-shelf 3D object detectors
to generate 3D region proposal candidates. Then we match candidate proposals with the encoded language query by a Matching Module. We find this simple pipeline achieves mediocre performance. Moreover, several limitations exist in this baseline. Firstly, it requires a separate 3D object detector and matching network, which leads to heavy computation cost. Secondly, the grounding performance largely relies on the quality of the 3D object detector, \textit{i.e.}, it is impossible to predict the correct region if none of the candidates access the targeted region. Therefore, a Multi-modal Single Shot Grounding (MSSG) model is further proposed to handle these problems. It jointly learns the detection model and the language model in an end-to-end fashion.
At inference time, it directly gives the final grounding result without any post-processing modules like Non-maximum Suppression (NMS). Such a cross-modal fusion mechanism enables the network to focus on the targeted region and largely improves the accuracy and efficiency compared with the baseline model.

Moreover, LiDAR point clouds usually contain rich geometric information about the surrounding environment, such as the position, shape, and motion of objects, etc., and images usually contain information about the appearance and texture of objects. Some approaches, such as DeepFusion~\cite{li2022deepfusion}, BEVFusion~\cite{liang2022bevfusion} and TransFusion~\cite{bai2022transfusion} use the complementary relationship between point clouds and images to obtain more comprehensive and accurate environmental perception. We also empirically find there is usually color information in natural language commands. Therefore, We further design a multi-modal fusion module to integrate the image feature into the point cloud representation, which can flexibly incorporate the information of the point cloud, image, or natural language.

To sum up, our main contributions are as follows:

\vspace{-1mm}
\begin{itemize}
    \item We propose a new multi-modal visual grounding task, named LiDAR Grounding, that aims to localize a 3D region given a natural language command. To the best of our knowledge, this is the first effort to incorporate natural language with 3D LiDAR point cloud and camera image. 
    \item A Multi-modal Single Shot Grounding (MSSG) model, with an effective and flexible token fusion method, is further proposed in a compact way by jointly learning the point cloud and language features. Furthermore, It also enables incorporating image feature to obtain rich semantic information.
    \item Experimental results on the Talk2Car dataset with elaborately designed evaluation metrics demonstrates the effectiveness of our proposed models. We expect this work presents a promising direction for the autonomous driving community.
\end{itemize}

\section{Related Work}

\textbf{3D Referring Expression Comprehension.}
The 3D Referring Expression Comprehension (REC) task, a.k.a. 3D Visual Grounding, aims to ground a natural language to a referred object in 3D point cloud space. 
Referit3D~\cite{achlioptas2020referit3d} proposes two datasets, termed as Sr3D (machine label) and Nr3D (human label). ScanRefer~\cite{chen2020scanrefer} is introduced similarly, but focus on grounding bounding box rather classify predicted ones. A lot of methods~\cite{huang2021text,chen2020scanrefer,zhao20213dvg,yuan2021instancerefer} have been proposed to tackle this task.
 A common pipeline is to first generate a bunch of 3D region proposals by 3D detector~\cite{charles2017deep,qi2019deep} or segmentation module~\cite{graham20183d}. Meanwhile the language expression is embedded by a Language Encoder~\cite{chung2014empirical}. Then the matching network links the 3D proposals to the language expression based on the similarity scores of features~\cite{chen2020scanrefer,yuan2021instancerefer}, graph neural network~\cite{huang2021text}, or Transformer~\cite{zhao20213dvg}. Finally, the proposal with the highest score will be output as the grounding result.
More recently,~\cite{yang2021sat} proposes to use 2D image semantics to assist 3D REC. However, all these approaches only focus on in-door scenes, which have difficulty in adapting to outdoor scenes. In this paper, we explore the 3D REC in an autonomous driving setting, called LiDAR Grounding. The more noisy and sparse LiDAR point clouds as well as the sophisticated commands pose new challenges to the grounding algorithms.

\noindent\textbf{2D Referring Expression Comprehension.}
The 2D Referring Expression Comprehension (REC) task~\cite{Jing_2021_CVPR,Luo_2020_CVPR,liang2021rethinking,wang2021structured,zhao2021cascaded,hui2021collaborative} has attracted a lot of attention and is widely discussed, which involves locating the targeted instance in 2D images queried by the given natural language expression. Most early methods~\cite{wang2019neighbourhood,Yang_2019_CVPR,Yu_2018_CVPR,wang2018learning} follow multi-stage pipeline. They typically first extract the region proposals from 2D object detectors~\cite{ren2015faster,Lin_2017_ICCV,he2016deep,duan2019centernet}, meanwhile abstracting the language features with a language encoder. After that, they rank the candidate proposals and adopt a multi-modal matching network to select the best matched proposal. However, such multi-stage methods usually encounter high computation cost and the upper bound of performance is largely determined by the 2D object detectors. To address these issues, Yang et al.~\cite{Yang_2019_ICCV} propose a fast one-stage paradigm that fuses a text query embedding into YOLOv3~\cite{redmon2018yolov3}, enabling an end-to-end optimization of the grounding model. Song et al.~\cite{Song_2021_CVPR} further explore the 2D REC task in video domain and propose semantic attention learning to disentangle the referring cues from the input language expression.
Built on the advances in 2D referring tasks, in this work, we tackle the LiDAR Grounding problem in the autonomous driving field.

\noindent\textbf{3D Object Detection.} 
Recently object detection in 3D domain has attracted a lot of research interest. We roughly categorize the approaches into two streams. Point-based methods~\cite{shi2020points,yang2019std,chen2019fast,yang20203dssd,shi2020pv} extract a set of point representation with the raw point cloud as input. Sampling and grouping~\cite{qi2017pointnet++,qi2017pointnet} techniques are often utilized. 3DSSD~\cite{yang20203dssd} also develops a fusion sampling strategy to enable high accuracy and speed.
Voxel-based~\cite{Zhou_2018_CVPR,yan2018second,yang2018pixor,simony2018complex,lang2019pointpillars,liu2020tanet} methods transforms the sparse points to dense voxel representation. To improve time efficiency, sparse convolution~\cite{yan2018second} is introduced. Voxel R-CNN~\cite{deng2021voxelrcnn} designs voxel ROI Pooling to take full advantage of multi-scale voxel features, achieving impressive performance. 
PV-RCNN~\cite{Shi_2020_CVPR} combines the advantages of both 3D voxel convolution and point-based abstraction, resulting a voxel-keypoint-grid pipeline, achieving remarkable gains. Recently, methods~\cite{bai2022transfusion,liang2022bevfusion,li2022deepfusion,liu2022bevfusion,li2022unifying,chen2022autoalignv2,yang2022deepinteraction,chen2022futr3d} based on multi-modal fusion have achieved better performance, since images can provide rich context information. Inspired by this, we also explore the effect of image features on the proposed task.

\section{Approach}

In this section, we first give the definition of the LiDAR Grounding task in~\ref{setup}, and then present the baseline method grounding-by-detection in~\ref{GBD}. The multi-modal single shot grounding model  is illustrated in~\ref{SSG}. We further present how to use image feature in~\ref{ssg-img}.

\subsection{Problem Setup}
\label{setup}
We define LiDAR Grounding as a new 3D referring expression comprehension (REC) task in an autonomous driving setting. It aims to localize a region of the LiDAR point cloud described by the natural language.
Formally, given a natural language expression $Q$ with $L$ words, $Q = \{w_1, w_2, \dots , w_L\}$ and a point cloud scene represented as a set of 3D points, $P = \{p_1, p_2, \dots , p_N\}$, where $N$ is the number of points and each $p_i$ is a vector of 3D location coordinates $(x, y, z)$ plus extra feature channels such as intensity measurements $r$.
The goal is to predict a 3D bounding box $B$ of the object referred by $Q$, where $B = (u, v, d, w, l, h, \alpha)$ consists of center location $(u, v, d)$ relative to ground plane and 3D size $(w, l, h)$ with rotation angle $\alpha$.

\subsection{Baseline}
\label{GBD}
\begin{figure}[t]
\centering
\includegraphics[scale=0.35]{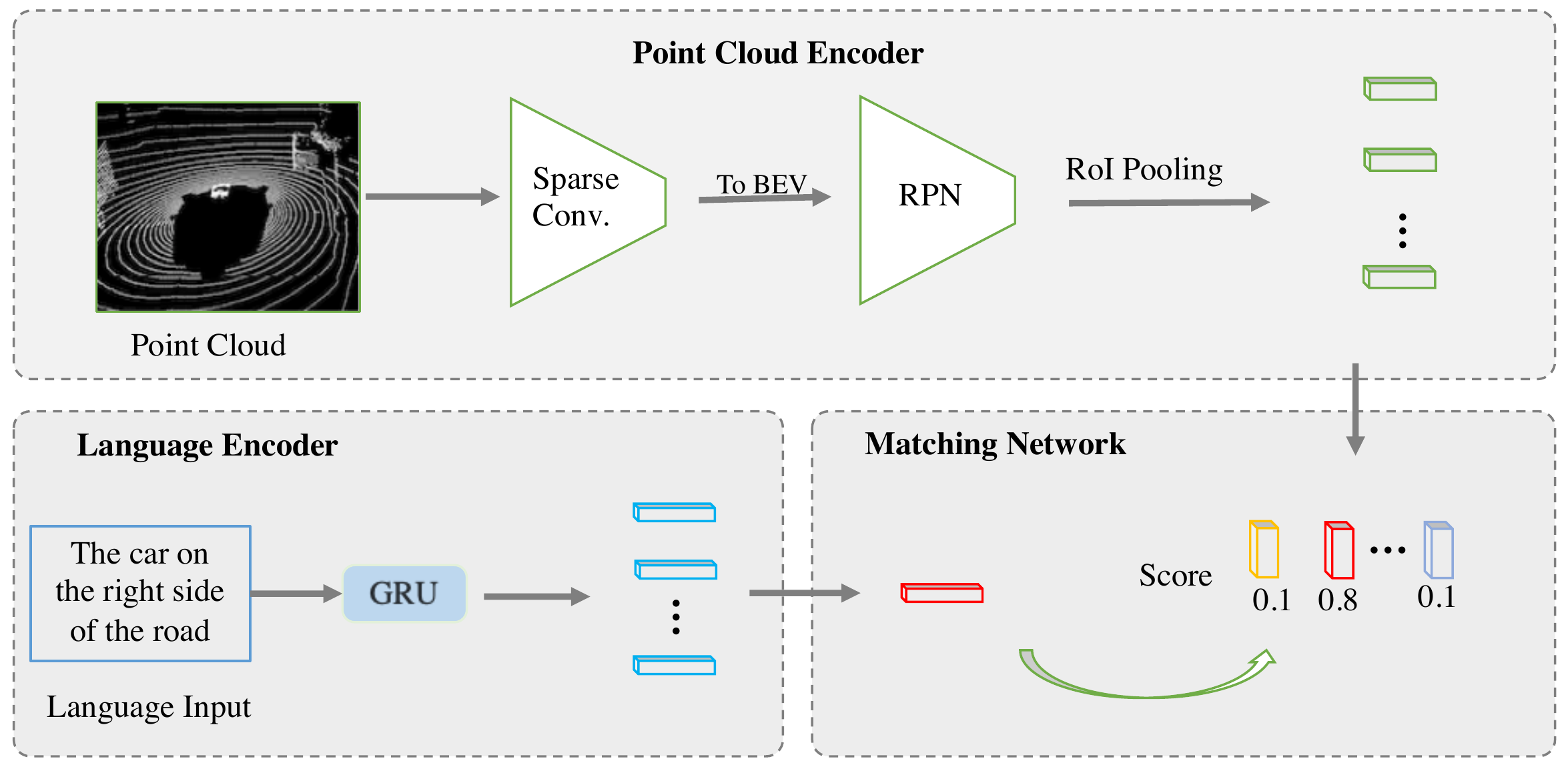}
\caption{Overview of the baseline model. Firstly the region proposals are extracted by a pre-trained 3D detector and processed by 3D RoI Pooling, and the language expression is encoded by a GRU. Then a Matching Network is utilized to match each candidate proposal with the language query. Finally, the one with highest score is selected as the final grounding result.}
\label{fig:two_stage}
\end{figure}

 Most existing methods~\cite{chen2020scanrefer,huang2021text,zhao20213dvg,yuan2021instancerefer} for indoor referring task can be summarized as a  grounding-by-detection (GBD) manner. 
Follow this pipeline, we firstly design a baseline for the proposed task. Fig.~\ref{fig:two_stage} illustrates the overview of our pipeline, which consists of two stages. Specifically, in the first stage, a bunch of 3D object region proposals are produced by an off-the-shelf 3D object detector. Then we extract the proposal features from the detector using 3D RoI pooling. In the second stage, the language expression is embedded by a Language Encoder with a GRU~\cite{chung2014empirical} or BERT~\cite{vaswani2017attention}. After that, the matching network links the 3D proposals to the language expression based on the similarity scores of features. Finally, the proposal with the highest similarity score will be output as the grounding result. Next, we will discuss each module in detail.

The 3D object detector plays a key role in the whole framework, since the final prediction is based on the candidate regions generated by it, and if the ground truth object is not covered, the subsequent work will be meaningless. Therefore, we employ a state-of-the-art 3D object detector CenterPoint~\cite{Yin_2021_CVPR} to extract the region proposals $P$:
\begin{equation}
    P = \{p_1, p_2, ... , p_K\}
\end{equation}
where $K$ is the number of proposals and each $p_i$ is  the bounding box of the candidate regions. Then, in order to extract the corresponding proposal features $F_{p_i}$, we attempt to fuse the features of center points along the five faces of a 3D bounding box following~\cite{Yin_2021_CVPR}:
\begin{equation}
    F_{p_i}= RoI (p_i), \quad i = 1, 2, ... , K
\end{equation}
where \textit{RoI} denotes 3D RoI pooling.

As for Language Encoder, we use a bi-directional GRU to extract the linguistic feature, and just summarize the last final hidden state in the forward and backward process to get the global language feature representation $F_t$.

In Matching Network, we first map the point feature and language feature into the latent space:
\begin{align}
    \hat{F}_{p} = f(F_{p}) \quad
    \hat{F}_t = f(F_{t})
 \end{align}
where $f$ means Multi-Layer Perceptrons (MLP) followed by L2 normalization to remap the feature value ranging from 0 to 1.

Finally, to measure the correlation or get the matching score between the features of each candidate region and language query, we have:
\begin{equation}
    score_{p_i} = cos(\hat{F}_{p_i}, \hat{F}_t), \quad i = 1, 2, ..., K.
\end{equation}
where $\textit{cos}$ denotes the cosine similarity function.

During training, we select the top $K$ region proposals based on the confidence score generated by the pre-trained detector. The proposal with highest intersect-over-union (IoU) with the ground truth box in the bird's eye view will be viewed as the positive, and other proposals are the negatives. The scenes that have all the candidate proposals with lower IOU between the ground truth box will be ignored. We freeze the Point Cloud Encoder and only update the parameters of Language Encoder and the Matching Module by minimizing the Cross Entropy Loss.
\begin{equation}
    L = -s^*\text{log}(s_i), \quad s_i = softmax_i(score_{p_i})
\end{equation}
where $s^*$ corresponds to the binary ground truth.
At inference time, the candidate proposal with the highest matching score will be selected as the final prediction.

\begin{figure*}[tpb]
\centering
    \includegraphics[scale=0.57]{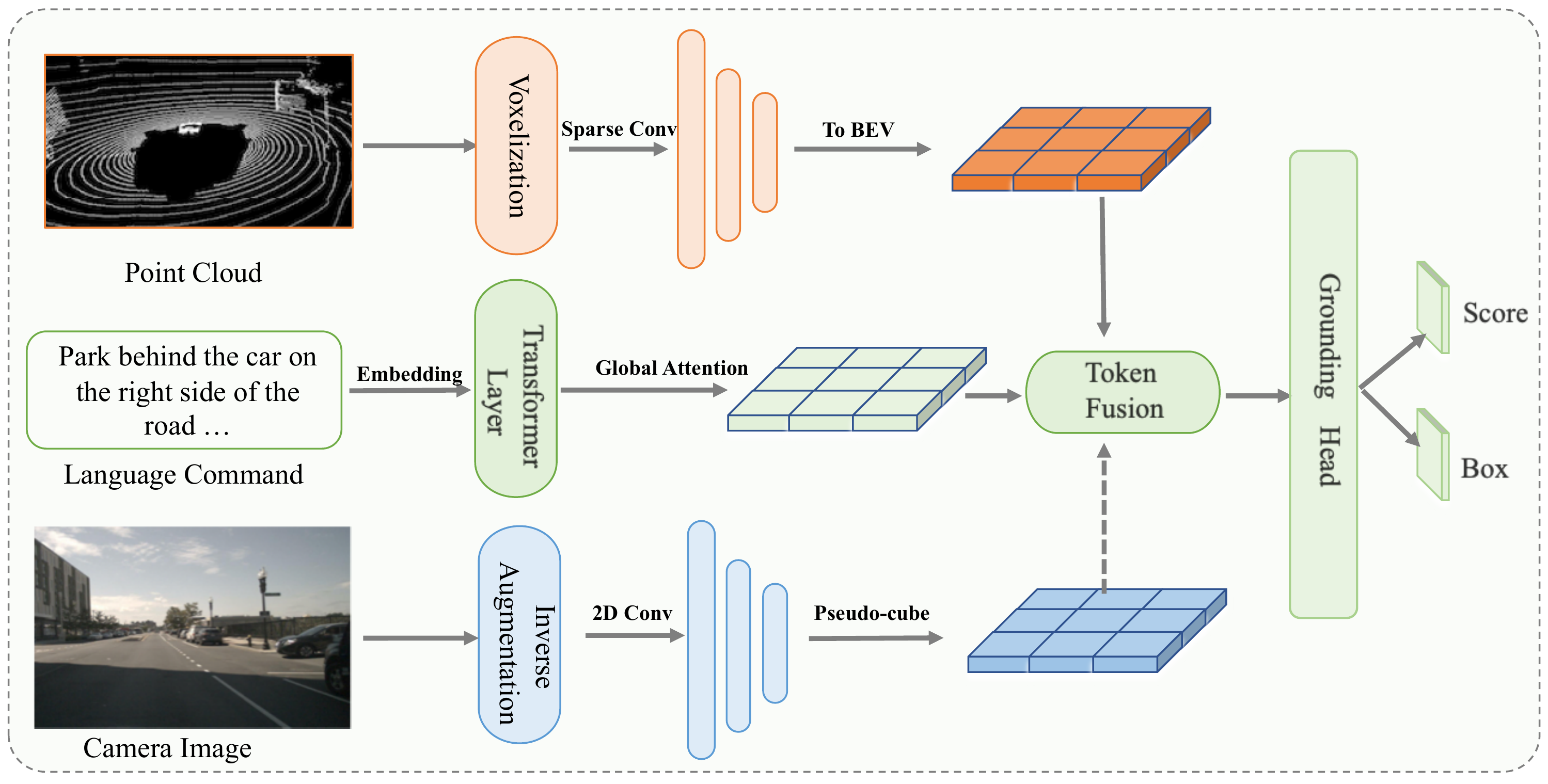}
    \caption{Overview of our multi-modal single shot grounding model. This method enforces cross-modal feature interaction and joint optimization between point cloud and language expression representations in a single shot fashion. The image feature could be optional input to provide semantic information.}
    \label{fig:one-stage}
\end{figure*}

\subsection{Multi-modal Single Shot Grounding}
\label{SSG}

According to the experimental results in Section~\ref{subsec:results}, the baseline model has achieved robust performance . However, there are still some limitations. First, the final performance is capped by the 3D object detector in the first stage. For example, the matching network will never give a correct answer if none of the predicted 3D region proposals have covered the targeted region. Second, the 3D object detector and matching network are separately learned, which lacks effective feature fusion and information exchange. To tackle these problems, we further present a single shot grounding (MSSG) model that enforces cross-modal feature interaction and end-to-end optimization for point cloud and language features. The pipeline is shown in Fig.~\ref{fig:one-stage}.

\noindent\textbf{Language Encoding.}
For the linguistic feature extraction, we try both BERT~\cite{vaswani2017attention} layer and bidirectional GRU~\cite{chung2014empirical} to show the importance of language encoding. Here we take the former case as example to illustrate our method. Formally, given the language expression $Q = [w_1, ... , w_L]$, we have:
\begin{equation}
    [u_1, \dots , u_L] = \text{BERT}(\hat{w}_1, \dots , \hat{w}_L) \label{eq:one_gru}
\end{equation}
where $\hat{w_j} = embed(w_j)$ is the embedding of $j\text{-}th$ word in the input expression with length $L$, and the output $u_j$ will contain context information. To get global feature representation, we adopt the pre-defined special token [CLS] as the fused text feature $f_{text}$.

\noindent\textbf{Point Encoding.}
To build a representation of point cloud, we firstly voxelize the input point cloud. Subsequently, sparse convolution is performed, and multi-scale voxel-wise features are aggregated to obtain a deep feature representation. Then the height and channel dimension will be concatenated to get the bird's eye view (BEV) features. In this way, we obtain the point cloud features $f_{point}$ of a scene.

\noindent\textbf{Cross-modal Fusion.}
Next, we combine the information from the point cloud branch and the language branch. We first map the language features and point features to a latent embedding space, then the element-wise product gives the cross-modal features. Formally, we have:
\begin{equation}
    f_x = (f_{text} W_t) \odot (g (f_{point})) \label{eq:fuse}
\end{equation}
where $W_t$ is learnable parameter, and $g$ represents $1\times1$ 2D convolution followed by batch normalization and ReLU. The language embedding will automatically broadcast to integrate into the point cloud feature space.  Due to the sparsity of 3D LiDAR point clouds, 2D convolutions are often utilized to enrich feature representation. Inspired by the success of vision transformers~\cite{xie2021segformer,liu2021swin,dosovitskiy2020image}, we adopt Window-based self Attention (WSA) to get a transformed feature. In each WSA block, we have:
\begin{align}
    U &= X_{i-1} + \text{Attn}(X_{i-1})  \notag \\
    X_{i} &= \text{LN}(U + \text{MLP}(\text{LN}(U)))  
\end{align}
where $X_{i-1}$ represents the output feature from previous layer, LN means layer normalization~\cite{ba2016layer}, Attn denotes the scaled dot-product attention~\cite{vaswani2017attention}, and MLP is the Multi-Layer Perceptron. Specially, in the MLP, we add a layer of 3x3 convolution to increase the receptive field.

\noindent\textbf{Grounding Head.}
Finally, the fused features $f_X$ are fed into a center-based head to predict the bounding box and confidence score. The general 3D object detector like Centerpoint~\cite{Yin_2021_CVPR} utilizes a K-channel heatmap head for each of $K$ classes to generate a heat map peak at the center of any detected object, optimized by Focal Loss~\cite{Lin_2017_ICCV}. However, in response to the proposed task, only one object instance provides positive supervised signal. The K-channel heatmap may contain a lot of redundant information and even impair the performance. Thus, we reduce the channel size into one to predict the confidence score of the referred object across all classes. As for the regression head, we use four separate sub-heads to predict the 2D center relative to the ground plane, 3D height, 3D size and rotation angle. 

\noindent\textbf{Negatives Sampling in Training.}
During the process of label assignment, only the location at ground truth object center will be assigned label 1, and the others will be labeled 0, since we only select one bounding box as the final result. However, the large spatial size of heat map \textit{i.e.}, $180\times180$ raises another issue. Different from 2D referring tasks~\cite{Jing_2021_CVPR,Yang_2019_CVPR,Song_2021_CVPR}, the image can be downsampled to small scales, such as $7\times7$ or $ 13\times13$. The large heatmap will bring too many negative samples, which may dominate the gradient propagation. To address this issue, we propose a binary mask map (BMM) to reduce the ratio of negatives. Only the locations with mask 1 will be taken into account for the gradients. Formally, the confidence score of location $(i, j)$ is calculated by:
\begin{equation}
    p(i, j) = \frac{b(i,j)~\text{exp}(h(i, j))}{\sum_{x,y}~b(x,y)~ \text{exp}(h(x, y))}
\end{equation}
where $b(x,y)$ is the value of binary mask map at location $(x,y)$ and $h(x,y)$ is the output logit confidence of heat map.

In our implementation, we use a simple but effective method to get BMM. Specifically, we randomly sample the negatives with probability $\gamma$. At each location,  it  has  probability $\gamma$ to be masked with 1. Particularly, it is guaranteed that the location of ground truth region center must be masked. We refer to the experiment part Section~\ref{subsec:results} for the analysis of BMM.

 \noindent\textbf{Training.}
During training, the confidence loss function is defined as follows:
\begin{equation}
    L_{conf} = - \sum_{i,j} g_{i,j} \text{log} (p_{i,j})
\end{equation}
where $g_{i,j}$ is the assigned label and $p_{i,j}$ is the output confidence score at location $(i,j)$. And the regression loss can be formulated by:
\begin{equation}
    L_{reg} = l (o, o^*)
\end{equation}
where $l$ represents L1 loss, and $o$ denotes the attributes of output bounding box at ground truth center $o^*$, including 3D center location, 3D size, sine and cosine value of yaw angle.

At last, we have the total training loss:
\begin{equation}
    L_{total} = L_{conf} + \lambda L_{reg}
\end{equation}
where $\lambda$ is the weighting coefficient.

\subsection{Multi-Modal Fusion}
\label{ssg-img}

Inspired by previous multi-modal fusion methods~\cite{bai2022transfusion,liang2022bevfusion,li2022deepfusion,liu2022bevfusion,li2022unifying,chen2022autoalignv2,yang2022deepinteraction,cheng2022learning}, we explore the effectiveness of image feature on the proposed task. Based on the fact that the object detection is very similar to the task proposed in this paper. The obvious difference is that the LiDAR Grounding only detects one candidate area, while the object detection task needs to detect all objects, but their shared features are similar. Therefore, an intuitive idea is that if the object detection task can benefit from image features, then the proposed task can also learn from this idea. In this way, the MSSG model can utilize the information of point cloud, natural language and camera image simultaneously.

\noindent\textbf{Token Fusion.}
We explore the scheme of fusion in the BEV space, that is to say, the features in the three modals of point cloud, language and image will be initially encoded by their own independent encoders, and then the information of each modal will be exchanged in the middle stage to carry out effective fusion. Adopting such a method has the following advantages. First, some pre-trained models can be used to introduce prior information to improve the performance of the model. Second, after the initial encoding by the encoders of the respective modalities, the feature representation of each modality is enriched, and a clearer understanding of its own meaning is obtained, and then in the process of interacting with information from other modalities, it is also easier to know what they need and what they can provide, thus lead to better fusion effect. 
The whole process can be termed as token fusion, and expressed as:
\begin{align}
     \hat{L} &= F_l(L) \notag \\
     \hat{P} &= F_p(P) \notag \\
     \hat{I} &= F_i(I) \notag \\
     X &= G([\hat{L}; \hat{P}; \hat{I}])
\end{align}
Where $L, P, I$ respectively represent the initial input language, point cloud and image features, and $F_l$ represents the language encoder. Recently, there are many language models based on Transformer that are pre-trained on large-scale datasets emerging. This approach can conveniently benefit from developments in the corresponding field. $F_p$ represents the encoder of the point cloud, here it can use the backbone network for 3D object detection with the best performance, or the pre-trained network on other datasets. $F_i$ represents the encoder of the picture, which can use either a traditional two-dimensional convolution-based network or a recent advanced Transformer-based model. $G$ represents a module that fuses the features of the three modalities. Here, it is represented by concatenating. In addition, point-by-point addition, multiplication, or other fusion methods using attention mechanisms can also be used. Then the fused feature $X$ can be used for downstream detection tasks.

In our implementation, we use a pre-trained ResNet-152~\cite{he2016deep} as the iamge encoder. After the image features are aligned with the point cloud, a pseudo-cube will be filled in as a pseudo-point cloud, which will be in one-to-one correspondence with the previous point cloud features to facilitate subsequent fusion operations in equation~\ref{eq:fuse}.

\noindent\textbf{Inverse Augmentation.}
During specific training process, data augmentation is often used to improve data efficiency, such as flip, rotation and translation. However, these transformation of point cloud is random and changes its original location in 3D space. It is necessary to align the image and point cloud features. Therefore, we record the augmentation details of point cloud and perform the same transformation in 2D image.

\section{Experimental Results}
\label{12}
\subsection{Dataset}
We evaluate the proposed methods on Talk2Car dataset~\cite{deruyttere2019talk2car} which is built upon the nuScenes dataset~\cite{Caesar_2020_CVPR}, a large-scale dataset for autonomous driving. The Talk2Car dataset contains 11,959 commands of 850 videos. Each command describes one target region in a frame and consists of 11.01 words, 2.32 nouns, 2.29 verbs and 0.62 adjectives on average. It has fine-grained classes the same as nuScenes, with 23 categories of different vehicles, pedestrians, mobility devices and others.  The dataset is split into train, validation and test sets with a number of 8,349, 1,163 and 2,447 commands, respectively. It has 3D bounding box annotations, but the box is annotated in the camera front view. Therefore, to get the box under point cloud, we convert the original box into LiDAR-top view using the corresponding calibrated parameters. Specifically, we firstly transform the box in camera view to ego view. Secondly, the ego vehicle frame for the timestamp of the image is converted into global plane. Thirdly, transform the global frame to ego view in LiDAR-top. Finally, the LiDAR box can be obtained according to the time step of the sweep. In each step, the translation and rotation angle will be transformed at the same time.
Since the test set is unavailable, we only conduct the experiments on the train and validation set.

\subsection{Evaluation Metrics}
We use the average precision (AP) as the evaluation metric. The match in BEV is defined by thresholding the Intersection-over-Union (IoU) between the predicted box and the ground truth in the bird eye view. The 3D AP further considers the accuracy of box height. The prediction is correct when the IoU is larger than a certain threshold. Due to the large scale variance of different object instance, we design two types of IoU threshold setting to evaluate the model performance.  Following the nuScenes~\cite{Caesar_2020_CVPR} detection task, we merge similar classes as well, resulting in 10 super classes. Specifically, the IoU threshold is presented in 
Table~\ref{tab:iou_thd}. Our motivation for setting two different types of thresholds is to show the accuracy of the box regression. Since in the process of inference, we only select the bounding box with the highest score as the final result. On the other hand, the volume of different objects in the dataset is quite different, and it is unreasonable to only set a unified threshold as the evaluation metric.



\begin{table}[t]
\renewcommand\tabcolsep{3.0pt}
\renewcommand\arraystretch{1.8}
\begin{center}
    \resizebox{0.99\columnwidth}{!}{
        \begin{tabular}{c c c c c c c c c c c}
        \hline
        Type & car& truck&c.v.& bus& trailer& barrier& motorcycle& bicycle& pedestrian& t.c. \\
        \hline
        A & 0.5 & 0.5 & 0.5 & 0.5 & 0.5 & 0.25 & 0.25 & 0.25 & 0.25 & 0.25 \\
        \hline
        B & 0.7 & 0.7 & 0.7 & 0.7 & 0.7 & 0.5 & 0.5 & 0.5 & 0.3 & 0.3 \\ 
        \hline
        \end{tabular}
    }
\end{center}
\caption{Two types of designed IoU setting. The c.v. means construction vehicle and t.c. denotes traffic cone. }
\label{tab:iou_thd}
\end{table}

\begin{table}[t]
\vspace{-5mm}
\renewcommand\arraystretch{1.5}
\renewcommand\tabcolsep{15.0pt}
\begin{center}
   \resizebox{0.99\columnwidth}{!}{
     \begin{tabular}{c c c c c}
     \toprule
     \multirow{2}{*}{Method} & \multicolumn{2}{c}{BEV AP} & \multicolumn{2}{c}{3D AP} \cr \cmidrule(r){2-3} \cmidrule(r){4-5}
     & @A & @B  & @A & @B \cr
     \midrule
     
     Talk2Car*~\cite{deruyttere2019talk2car}  & 30.6	& 24.4	& 27.9	& 19.1 \\ \hline
     MSSG (Ours)  & 37.8	& 26.1	& 31.9 & 20.3\\ \hline
     MSSG + Pre (Ours) & 48.0 & 34.1 & 43.0 & 22.4  \\ \hline
     MSSG + Pre + Img (Ours) & \textbf{50.1} & \textbf{35.7} & \textbf{45.4} & \textbf{23.7} \\ 

     \bottomrule
     \end{tabular}
   }
\end{center}
\caption{\textbf{Comparison of performance with different methods on Talk2Car~\cite{deruyttere2019talk2car} Validation Set.} Pre represents pre-training, Img means image features, and * denotes our re-implementation.}
\label{tab:overall_performance}
\vspace{-5mm}
\end{table}

\begin{table}[t]
\renewcommand\arraystretch{1.5}
\renewcommand\tabcolsep{12.0pt}
\begin{center}
  \resizebox{0.99\columnwidth}{!}{
    \begin{tabular}{c c c c c c c}
    \toprule
    \multirow{2}{*}{IoU Type} &
    \multicolumn{3}{c}{Training} &\multicolumn{3}{c}{Validation}\cr
     \cmidrule(r){2-4}  \cmidrule(r){5-7}
    & Found  & Total & Ratio & Found  & Total & Ratio \cr
    \midrule

    A & 5058 & 8328 & 0.61 & 643 & 1128 & 0.57 \\
    \hline
    B & 4056 & 8328 & 0.49 & 523 & 1128 & 0.46 \\
    \bottomrule
    \end{tabular}
  }
\end{center}
\caption{Statistics of the proportion that covers ground truth in proposals generated by pre-trained CenterPoint~\cite{Yin_2021_CVPR}. }
\label{tab:upper_bound}

\end{table}

\subsection{Implementation Details}
For the baseline method, we extend the Talk2Car~\cite{deruyttere2019talk2car} approach in a 3D manner. Specially, in terms of the Point Cloud Encoder, we pre-train a two-stage CenterPoint~\cite{Yin_2021_CVPR}  with VoxelNet~\cite{Zhou_2018_CVPR} as backbone on nuScenes~\cite{Caesar_2020_CVPR} following original setting~\cite{Yin_2021_CVPR}. The number of selected region proposals is set to $K=128$. The Language Encoder uses Bi-GRU with a 512 hidden size.  In Matching Network, the dimension of latent space is set to 512. We train the model for 20 epochs with a batch size of 4. The stochastic gradient descent (SGD) is used with momentum 0.9, weight decay 1e-4 and initial learning rate 0.01 with a drop rate of 0.1 at 16-th epoch.

For the proposed single shot grounding (MSSG) approach, the window size in WSA block is set to 9, and we use 2 blocks for time efficiency. For the language encoder, we just use 2 layers of BERT for a relatively fail comparison.  Since the ground truth regions only appear in the camera front view, we filter the point clouds, leaving only those that can be projected into the image of the camera front sensor. We only use global scaling and translation as the augmentation method. The weighting coefficient $\lambda$ is set to 0.25.
In terms of the training details,  we use Adam~\cite{Kingma2015adam} optimizer to train the model for 20 epochs with bach size 4. We use one-cycle learning rate policy~\cite{Sylvain_one_cycle}, with max learning rate 1e-3, weight decay 0.01, and momentum from 0.85 to 0.95. When utilizing pre-training mwthod, we design a simple detection head for MSSG, then we train the new model on NuScenes training set to acquire some prior knowledge The pre-trianed network weights are used to initialize the backbone of the original MSSG.

\subsection{Results}

\begin{table}[t]
\renewcommand\arraystretch{1.5}
\renewcommand\tabcolsep{16.0pt}
\begin{center}
   \resizebox{0.99\columnwidth}{!}{
     \begin{tabular}{c c c c c c}
     \toprule
     \multirow{2}{*}{Model} & \multicolumn{3}{c}{Component} & \multirow{2}{*}{NDS} & \multirow{2}{*}{mAP}\cr
       \cmidrule(r){2-4} &
     Base & Img & Aug \cr
     \midrule
     1 & \checkmark & & & 43.4 & 32.6 \\ \hline
     2 &  \checkmark  & \checkmark & & 44.0	& 34.2  \\ 
     \midrule
     3 &  \checkmark & & \checkmark & 47.5 & 41.8 \\ \hline
     4 & \checkmark & \checkmark & \checkmark& 47.6 & 42.2 \\
     \bottomrule
     \end{tabular}
   }
\end{center}

\caption{\textbf{Ablation study showing the effectiveness of image feature in object detection on 1/4 NuScenes~\cite{Caesar_2020_CVPR} dataset.} Img represents image feature, and Aug means data augmentation.}
\label{tab:ablation_study_1_4_det}

\end{table}

\begin{figure}[!t]
      \centering
      \includegraphics[scale=0.30]{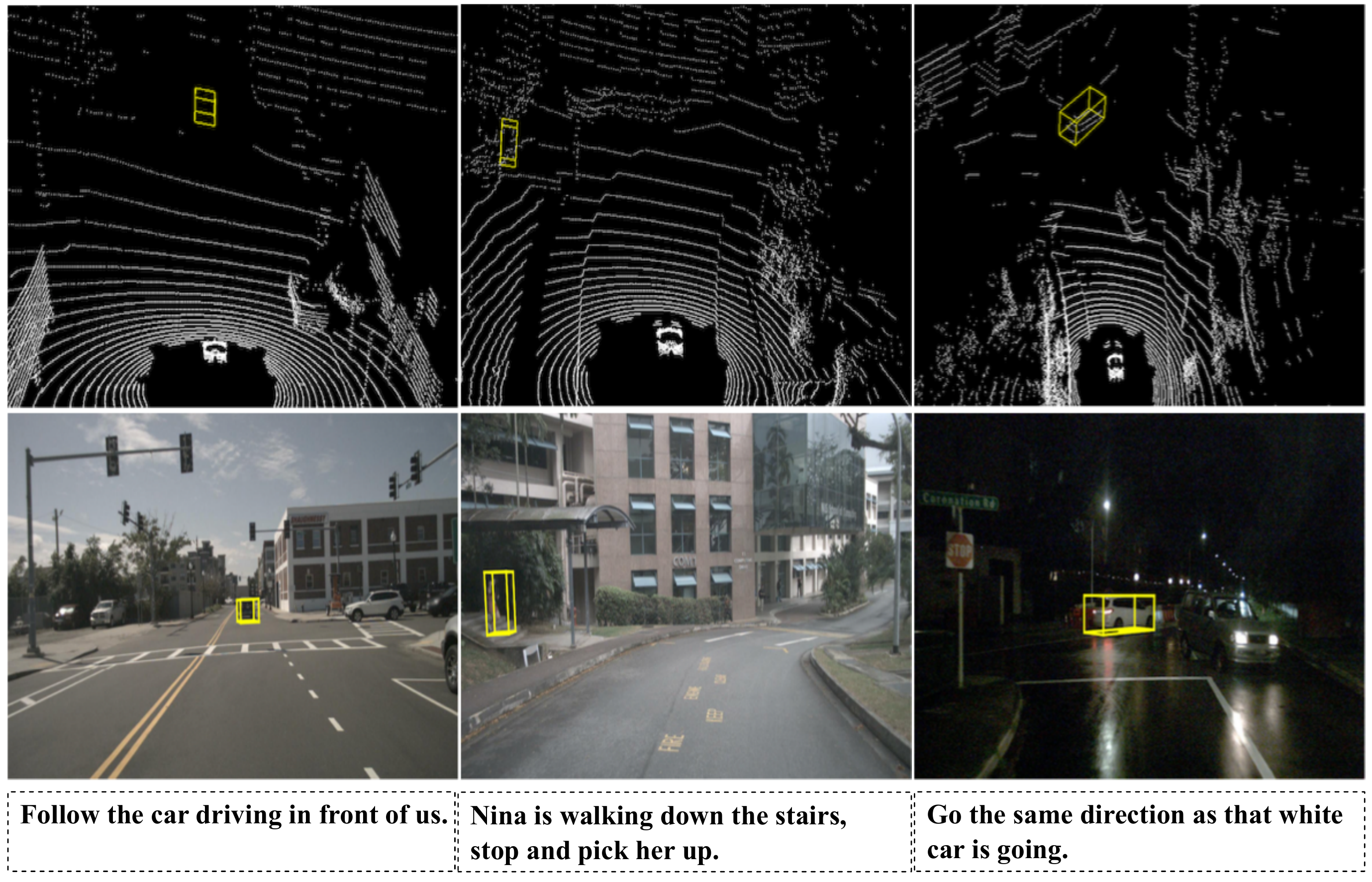}
      \caption{Three correct examples inferred by MSSG on Talk2Car validation set. Grounded 3D bounding boxes are shown in yellow. }
      \label{fig:vis_corr}
 \end{figure}

\textbf{Overall Performance.} Table~\ref{tab:overall_performance} shows the overall performance of the proposed methods on Talk2Car~\cite{deruyttere2019talk2car} validation set. We report the average precision (AP) over all classes under two types of IoU setting, where @A means \textit{Type-A} and @B represents \textit{Type-B}. From the table, we can find that the single shot grounding (MSSG) method significantly surpasses the grounding-by-detection (baseline) method. Our MSSG surpassed the baseline by 7.2\% under type-A setting in BEV space, 4.0\% in 3D space, which demonstrates the significant effectiveness of rich cross-modal feature interaction of MSSG pipeline. Moreover, when the MSSG adopts the pre-trained backbone, the performance gets significant improvement, \textit{i.e.}, 10.2\% in BEV-AP@A  and 11.1 \% of 3D-AP@A. Since the setting B that has a higher IoU threshold, there is an interesting phenomenon. The boost in condition B is relatively small compared with A. In other words, the baseline is more robust to evaluation metrics. This can be attributed to the off-the-shelf 3D detector. Since the quality of the region proposal is fixed, it is not sensitive to different IoU thresholds. More generally, it heavily depends on the performance of the pre-trained detector, but the MSSG approach learns feature representation from scratch, and the performance of the regression head may be slightly weak, resulting in inaccurate 3D box prediction.

\noindent\textbf{Upper-bound of Baseline Method.} Table \ref{tab:upper_bound} shows the ratio of positive bounding box in the proposals generated by the pre-trained 3D detector CenterPoint, i.e., exists a proposal which the IoU between the predicted box and the ground truth object is larger than the corresponding threshold. To a certain degree, the ratio can represent the upper-bound of the performance of the baseline method. Since if none of the candidates could cover the ground truth area, then it is impossible that the correct region will be ranked first in the second stage. Although the baseline has acceptable result, it helplessly has a clear upper bound on performance. To improve its performance in the future, maybe a more effective method is to consider enhancing the performance of the first-stage 3D detector.

\begin{figure}[t]
      \centering
      \includegraphics[scale=0.30]{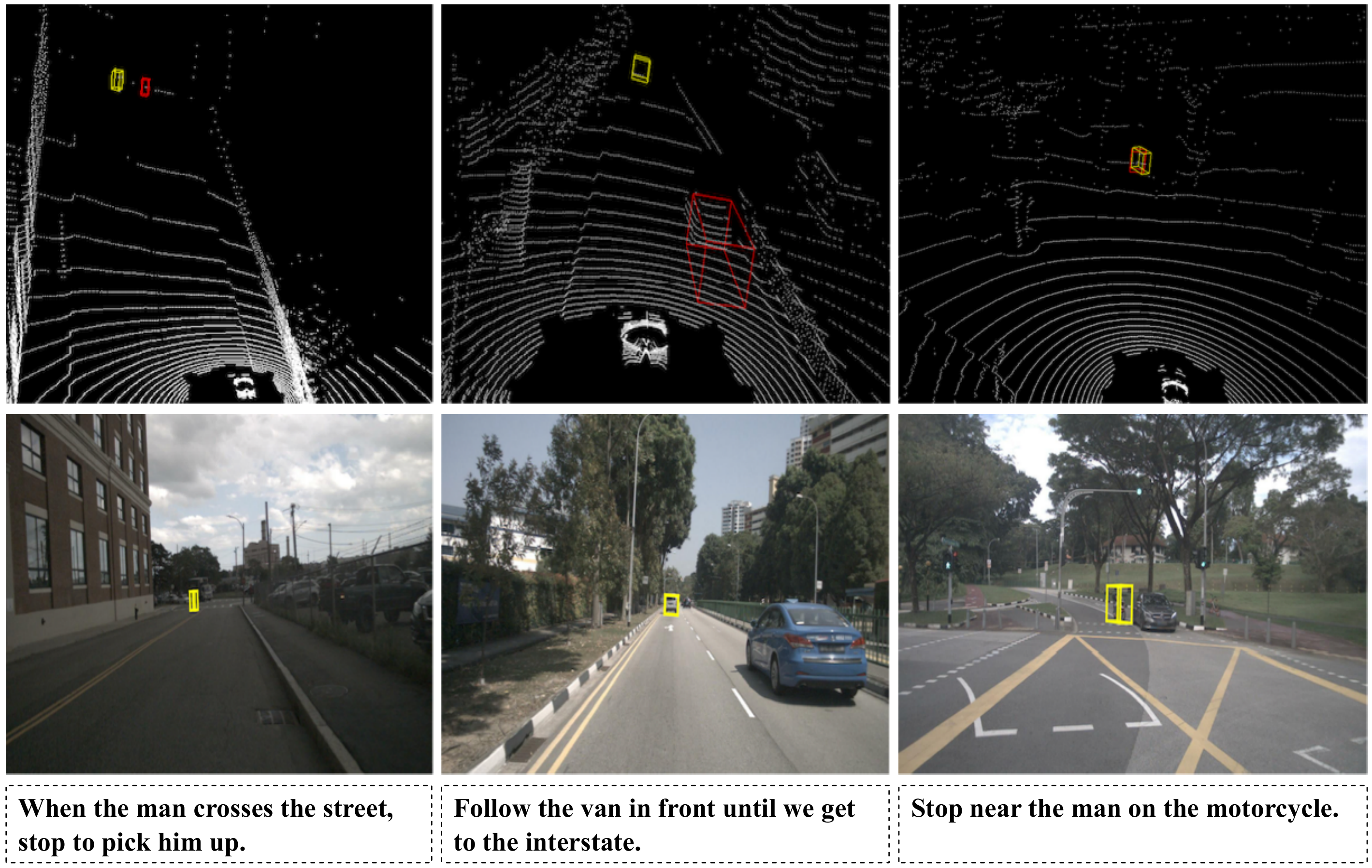}
      \caption{Three failed cases inferred by MSSG on Talk2Car validation set. The wrong predictions are shown in red color, while the ground truth boxes are visualized in yellow.}
      \label{fig:vis_fail}
\end{figure}

\noindent\textbf{Negatives Sampling.} Directly predicting the bounding box score at each location of large scale feature map will bring too many negative samples. To address this issue, we propose a binary mask map to sample some negatives. At each location, it has probability $\gamma$ to  be set to one. To reveal the effectiveness of $\gamma$, we have done sensitivity analysis with different sampling ratio on validation set. In our experiment, we  find when the sampling ratio ranges from 0.1 to 0.4, the performance will be better than those from 0.5 to 1.0. This indicates that the model performance can benefit from reducing the number of negative examples appropriately. However, if the number of sampled negative examples is particularly small, the gradients at many positions can not be back-propagated, which may disrupt the training process. This can be demonstrated from that the best performance is achieved when the sampling ratio equals 0.4.

\noindent\textbf{Effectiveness of Image Feature.} To enhance the interaction of cross-modal features, we further fuse the point cloud and image features. To reveal the effectiveness of image feature, we firstly conduct a pilot experiment. As shown in Table~\ref{tab:ablation_study_1_4_det},  we select the Centerpoint as the baseline, trained the model on 1/4 nuScenes dataset for time efficiency. When compare model \#1 and \#2, the NDS is improved 0.6\% and mAP is improved 1.6\%, which indicates the image feature can enrich the point cloud representation. In practical training setting, data augmentation is often used to improve data efficiency. However, the transformation of point cloud is random and changes its original location in 3D space. It is necessary to align the image and point features. Therefore, we record the augmentation details of point cloud and perform the same transformation in 2D image. From the table, we can find the performance still have gains when compared model \#3 with \#4. The pilot experiment shows the believable benefit of image feature, then we apply the same fusion method in our MSSG model. As shown in Table~\ref{tab:overall_performance}, we can find the image feature also boosts a lot, which indicates that the model's understanding of location information can be strengthened by deep cross-modal feature fusion. This also indicates the researchers can  explore more powerful fusion module in future work.

\noindent\textbf{Qualitative Results}
Fig.~\ref{fig:vis_corr} depicts the visualization results of single shot grounding model on Talk2Car validation set, which shows the impressive results of our method. Nevertheless, the overall performance of our method is still not high (\~50\% BEV-AP@A), so we also visualize some bad cases  in Fig.~\ref{fig:vis_fail} to ask for directions that can improve the performance. For the first failed case, long-distance localization is the major difficulty. The model has inferred that the pedestrian needs to be located, but the long-distance target is too small to be successfully located. For the second case, the interference of multiple targets should be the most important issue. In other words, the model does not understand the language query well. As discussed before, the model tends to predict salient objects. Therefore, how to enhance the model's anti-interference is an important point. For the third case, the model is only one step away from success. Unfortunately the location predictions are not accurate. This indicates that the regression head of the model needs to be strengthened. This also reflects the benefits of setting two different types of thresholds from the side.
Overall, we hope there is progress and improvement for long-distance targets, semantic understanding, and precise location regression in future work.

\label{subsec:results}

\section{Conclusion}
In this paper, we proposed a new task in the autonomous driving scenario, \textit{i.e.}, LiDAR Grounding, that aims to ground a natural language command to the referred region in LiDAR point clouds. To tackle this task, we firstly summarize previous efforts into a new baseline, grounding-by-detection model. This method follows a two-stage pipeline: first detection and then grounding. It extracts language features and region proposal features separately, and selects the grounding result according to the similarity score generated by another matching network. We also propose an end-to-end method, sing shot grounding (MSSG) model. It enables end-to-end optimization and is more fast and accurate, thanks to the cross-modal feature interaction and fusion. We also explore the effectiveness of image feature, which has rich semantic information, leading to better performance.  Extensive experiments on the Talk2Car dataset show solid results. We hope our work could present a promising direction in the autonomous driving community.



\bibliographystyle{ACM-Reference-Format}
\bibliography{sample-sigconf.bib}

\end{document}